\pdfoutput=1
\documentclass{egpubl}
\usepackage{pg2022}

\SpecialIssuePaper         
\usepackage{verbatim}
\usepackage[T1]{fontenc}
\usepackage{dfadobe}  
\usepackage{multirow}
\usepackage{cite}  
\usepackage{color}
\usepackage{amsmath,amssymb}
\usepackage{hyperref}
\hypersetup{hidelinks,
	colorlinks=true,
	allcolors=blue,
	pdfstartview=Fit,
	breaklinks=true}
\BibtexOrBiblatex
\electronicVersion
\ifpdf \usepackage[pdftex]{graphicx} \pdfcompresslevel=9
\else \usepackage[dvips]{graphicx} \fi

\usepackage{egweblnk} 

\title[Contrastive Semantic-Guided Image Smoothing Network]%
{Contrastive Semantic-Guided Image Smoothing Network}

\author[J. Wang et al.]
{\parbox{\textwidth}{\centering Jie Wang$^{1,2}$,
Yongzhen Wang$^{1}$, 
Yidan Feng$^3$, 
Lina Gong$^1$, 
Xuefeng Yan$^1$, 
Haoran Xie$^{4\dag}$, 
Fu Lee Wang$^{5}$, 
Mingqiang Wei$^{1,2}$\thanks{Co-corresponding authors: H. Xie and M. Wei.}
	}
	\\
	{\parbox{\textwidth}{\centering $^1$School of Computer Science and Technology, Nanjing University of Aeronautics and Astronautics, Nanjing, China\\
			$^2$Shenzhen Research Institute, Nanjing University of Aeronautics and Astronautics, Shenzhen, China\\
			$^3$Centre for Smart Health, School of Nursing, Hong Kong Polytechnic University, Hong Kong, China \\
			$^4$Department of Computing and Decision Sciences, Lingnan University, Hong Kong, China\\
			$^5$School of Science and Technology, Hong Kong Metropolitan University, Hong Kong, China
		}
	}
}

\begin{document}
	\maketitle
	\begin{abstract}
	
		Image smoothing is a fundamental low-level vision task that aims to preserve salient structures of an image while removing insignificant details. Deep learning has been explored in image smoothing to deal with the complex entanglement of semantic structures and trivial details. However, current methods neglect two important facts in smoothing: 1) naive pixel-level regression supervised by the limited number of high-quality smoothing ground-truth could lead to domain shift and cause generalization problems towards real-world images; 2) texture appearance is closely related to object semantics, so that image smoothing requires awareness of semantic difference to apply adaptive smoothing strengths. To address these issues, we propose a novel Contrastive Semantic-Guided Image Smoothing Network (CSGIS-Net) that combines both contrastive prior and semantic prior to facilitate robust image smoothing. The supervision signal is augmented by leveraging undesired smoothing effects as negative teachers, and by incorporating segmentation tasks to encourage semantic distinctiveness. To realize the proposed network, we also enrich the original VOC dataset with texture enhancement and smoothing labels, namely VOC-smooth, which first bridges image smoothing and semantic segmentation. Extensive experiments demonstrate that the proposed CSGIS-Net outperforms state-of-the-art algorithms by a large margin. Code and dataset are available at \href{https://github.com/wangjie6866/CSGIS-Net}{https://github.com/wangjie6866/CSGIS-Net}.

		\ccsdesc[300]{Computing methodologies~Image processing}
		
		\printccsdesc   
	\end{abstract}  
	\section{Introduction}
	Image smoothing is a fundamental technique widely used in various computer vision and graphics tasks, aiming to preserve salient structures while smoothing out insignificant details. Numerous algorithms have been proposed and applied to many downstream tasks such as texture removal \cite{xu2012structure,sun2017image}, detail enhancement \cite{fattal2007multiscale}, and clip-art restoration \cite{wang2006deringing, xu2011image}, which benefit from well-smoothed images. However, due to the complicated entanglement of structures and trivial details in natural images, image smoothing is still a challenging task that has attracted considerable attention in computer vision and computer graphics communities. Previous prior-based methods can be mainly divided into two classes: kernel-based and optimization-based methods. The kernel-based methods (e.g., Bilateral Filter \cite{tomasi1998bilateral} and Guided Filter \cite{he2012guided}) utilize local image statistics to explore the effective kernels, which may introduce inconsistent smoothing effects in different local areas. The optimization-based methods iteratively optimize the smoothing result under the global regularization terms, which trade smoothing quality with time and memory. These methods depend on manually designed image priors that describe low-level image features, which suffer from tedious parameter tuning and can be vulnerable to special input cases.

	With the development of deep learning, data-driven approaches \cite{xu2015deep,zhu2019benchmark} have also been explored on image smoothing. The deep hierarchical network structure has a natural advantage in capturing semantic cues for extracting meaningful image structures. However, the supervision signal for the smoothing task is hard to define. \cite{xu2011image} only approximates the smoothing results from existing operators, and \cite{lu2018deep} blends cartoon images with textures, which can cause over/under-smoothing on natural images due to the domain shift problem. These methods handle easy cases (salient structures and weak details) with better efficiency, but fail to tackle the coexistence of weak semantic structures and strong trivial details. Differently, \cite{feng2021easy2hard,zhu2019benchmark} build high-quality ground-truth domains by delicately designed screening procedures, such as comprehensive user studies and patch-level quality measurements. Although achieved notable gain in visual performance, there are still two ignored problems: 1) since the desired smoothing ground-truths are hard to collect, current methods are biased to the training domain when only supervised by the limited number of ground-truth data. 2) Current methods are dominated by pixel-wise constraints, which encourage smooth recovery of the overall structure, but neglect the fact that texture details are closely related to object semantics. For instance, the textures of grass and wall are totally different, and thus require different smoothing strengths.
	 
	To resolve the aforementioned problems, we propose to introduce contrastive learning and semantic guidance into the image smoothing problem. For the first problem, we augment the supervision signal by the undesired smoothing effects (i.e. over/under-smoothed images) as negative samples, where the predicted results are pushed away from those negative samples in a contrastive manner to encourage more robust representation learning for real-world image smoothing. For the second problem, we enhance the semantic awareness of the smoothing network by incorporating edge detection and semantic segmentation as auxiliary tasks. Edge detection emphasizes the global semantics and helps to localize the meaningful structure areas to be preserved. Semantic segmentation requires distinctiveness in different types of objects and consistency within the same type, and thus encourages the smoothing network to be semantic adaptive. Also, to support the supervision of the semantic branch, we build a dataset named VOC-smooth based on PASCAL VOC 2012 \cite{everingham2015pascal} with both smoothing and segmentation labels, which makes the first trial to bridge image smoothing and semantic segmentation.
	
	In summary, our main contributions are as follows:
	\begin{itemize} 
		\item We propose a contrastive semantic-guided image smoothing network named CSGIS-Net that benefits from a contrastive learning paradigm and high-level semantic information, thereby enhancing the smoothing capacity of the model.
		\item We propose to build the VOC-smooth dataset, which contains 15,000 texture-enhanced images with both high-quality smoothing ground-truths and semantic segmentation labels.
		\item Extensive experiments demonstrate that the proposed CSGIS-Net performs favorably against the state-of-the-art image smoothing approaches.
	\end{itemize}
	 
	\section{Related work}
	
	\subsection{Image Smoothing Methods} 
	\textbf{Kernel-based methods:} Kernel-based approaches mainly apply the local image statistics to explore the effective kernels. For edge-preserving smoothing, Anisotropic Diffusion (AD) \cite{perona1990scale} defines a new scale-space and achieves a diffusion process iteratively. Bilateral Filter \cite{tomasi1998bilateral} has been widely used in various tasks, weighted by the Gaussian in both spatial and intensity distance. Gastal et al. propose the adaptive mainfold filter(AMF) \cite{gastal2012adaptive} to handle high-dimensional data as an extension to BF. They further introduce Domain Transform filtering (DT) \cite{gastal2011domain} in real-time. Lee et al. \cite{lee2017structure} propose a gradient-guided algorithm and adaptively smooth image gradients via the interval gradient operator. However, the aforementioned local filters often result in gradient reversals and artifacts(such as halos) \cite{farbman2008edge}. 
	
	\textbf{Optimization-based methods:} To address the issue caused by the local filters, optimization-based approaches are designed, which attempt to solve a global objective function consisting of a data term and a regularization term. TV regularization \cite{rudin1992nonlinear} and its variant RTV (relative total variation) \cite{xu2012structure} aim at preserving edges and the latter focuses on texture. The weighted least square smoothing (WLS) \cite{farbman2008edge} is proposed for progressive coarsening of images and multi-scale detail extraction. \cite{bhat2010gradientshop} presents an optimization framework for image processing (e.g. smoothing), which mainly utilizes the gradient-domain analysis. Furthermore, Ham et al. solve a non-convex optimization problem, by combining a static guidance weight with Welsch's penalty \cite{holland1977robust}, namely SD filter (SDF) \cite{ham2017robust}. The latest work \cite{liu2021generalized} introduces a truncated Huber penalty function and establishes a generalized framework, which is capable to achieve diverse smoothing natures under different parameter settings.
	
	\textbf{Data-driven methods:}
	Recently, many deep learning-based methods have been proposed and achieved remarkable performance. Xu et al. \cite{xu2015deep} propose a deep edge-aware filter to approximate different operators based on a gradient domain training procedure. Shen et al. \cite{shen2017convolutional} present a pyramid structure with an extended receptive field. In contrast, Zhu et al. \cite{zhu2019benchmark} achieve an end-to-end architecture utilizing the ResNet \cite{he2016deep} and VDCNN \cite{conneau2016very} as backbone. However, most of them fail to generate a novel smoothing effect since they are only supervised by the results of existing methods. Therefore, Fan et al. \cite{fan2018image} present an unsupervised learning paradigm to learn an energy function. More recently, Fan et al. \cite{fan2017generic} and Feng et al. \cite{feng2021easy2hard} combine the edge map, the former transfers edge prediction to a cascaded convolutional network, and the latter proposes a joint edge detection and image smoothing network.
	
	\subsection{Image Smoothing Benchmark}
	 Several datasets such as BSDS300 \cite{martin2001database}, DIV2K \cite{agustsson2017ntire}, MIT5K \cite{bychkovsky2011learning} have been widely used in existing algorithms. Unfortunately, due to the lack of corresponding ground-truth images for smoothing, the evaluation is quite subjective. To address this issue, Zhu et al. \cite{zhu2019benchmark} generate a set of ground-truths through traditional operators. Xu et al. \cite{xu2020pixel} build a synthetic dataset(NKS) by blending cartoon images and common textures. However, they still fail to tackle the real-world data since the highly biased ground-truths. For evaluation, pixel-wised Peak-Signal-to-Noise-Ratio (PSNR) \cite{wang2004image} and image-wised Structural Similarity index (SSIM) \cite{wang2004image} are often employed to measure the difference between the reconstructed image and the ground-truth.
	
	\begin{figure*}[!htb]
		\centering
		\includegraphics[width=1\textwidth]{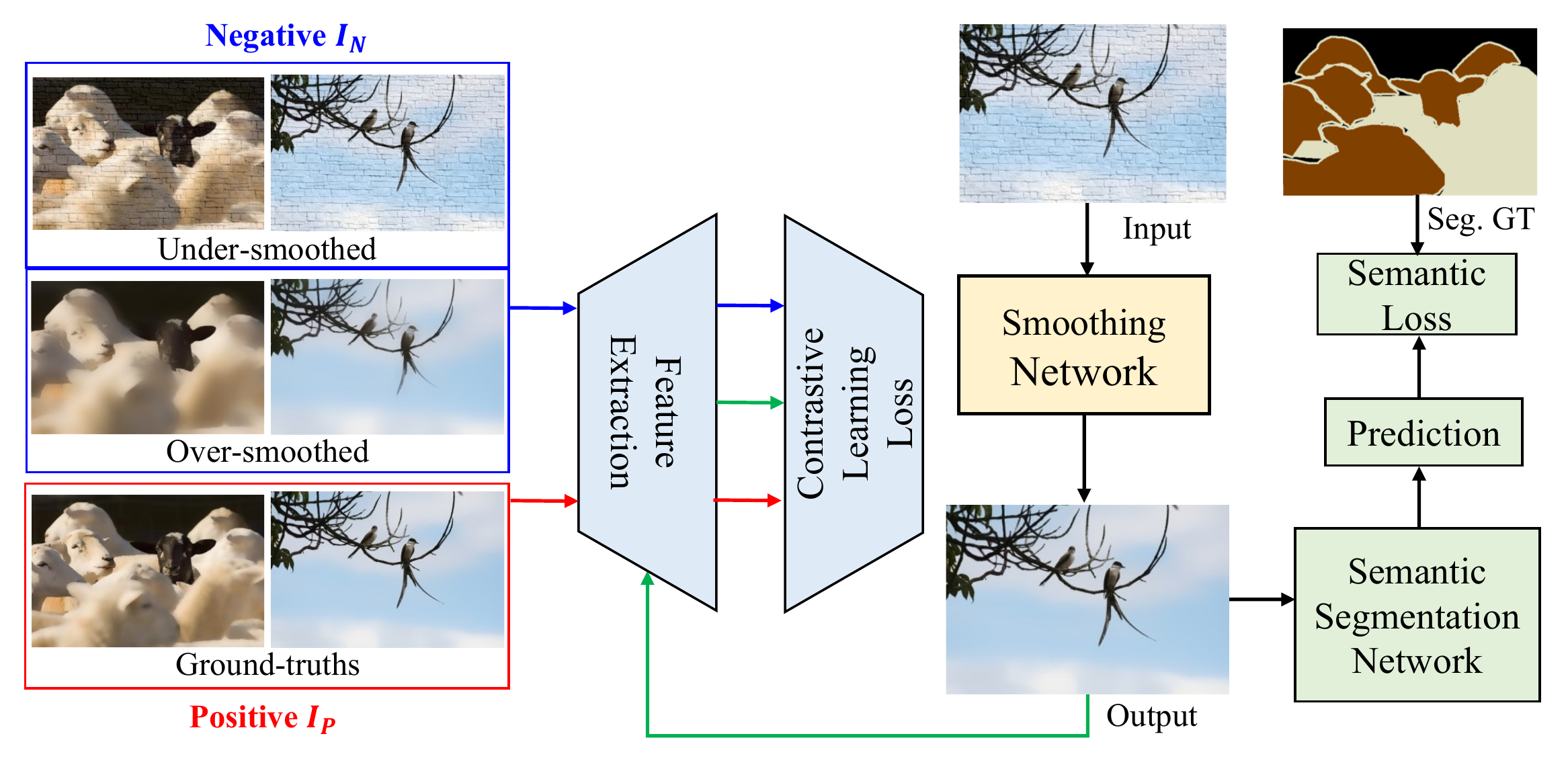}			
		\caption{Pipeline of our CSGIS-Net. It includes a smoothing network, a semantic segmentation network and a contrastive module.}
		\label{net1}
	\end{figure*}
	
	\subsection{Contrastive Learning}
	Contrastive learning, fully exploring images' supervisory information, has been widely adopted in self-supervised learning \cite{henaff2020data,tian2020contrastive}. For a given anchor, contrastive learning attempts to pull it together with positive pairs and push it far away from the negative ones. Previous works have realized such an idea in various computer vision tasks and achieved notable performance gains. The Noise Contrast Estimation(NCE) \cite{gutmann2010noise}, InfoNCE \cite{van2018representation}, and triplet loss \cite{hermans2017defense} are usually adopted as loss functions. Recently, \cite{park2020contrastive} demonstrates the effectiveness of contrastive learning for image reconstruction. However, few studies have focused on image smoothing since it's hard to define the positive/negative samples.

	\section{Proposed Method}
	In real-world smoothing scenes, strong details and weak structures of images are often intertwined, making the image smoothing task more challenging. Existing smoothing efforts are prone to cause over/under-smoothed problems since these algorithms are unable to understand the smoothing operation in a semantic manner. To address this issue, We regard the high-level semantic information as important prior knowledge to help the model better distinguish the texture and structure features. Meanwhile, the contrastive learning paradigm is adopted to augment the supervision signal by the unexplored information behind the negative samples (i.e. over/under-smoothed images), preventing the model's output from falling into over/under-smoothed situations. To this end, a novel contrastive semantic-guided image smoothing network (CSGIS-Net) is proposed that leverages both contrastive learning paradigm and semantic knowledge for image smoothing.
	
We first present the overview of the proposed CSGIS-Net (see Fig. \ref{net1}). Then the contrastive constraint and semantic structure preservation modules are described to exhibit how they work in our network. Finally, the loss function used in our training phase is introduced.

 	\subsection{Overview}
 	Fundamentally, the image smoothing task can be described as seeking a mapping function $\varphi(\cdot)$ from $I$ to $O$, where $O=\varphi(I)$. It is an ill-posed problem since the smooth ground-truth is hard to define. In our work, to facilitate robust image smoothing, we explore the data's own supervision information from two different priors: one is from the contrastive learning paradigm, which introduces the negative samples $I_{N}$ and positive ones $I_{P}$ to alleviate the domain shift. The other is the semantic guidance information, i.e. the ground-truth and the prediction of semantic segmentation ($S_{gt},S_{pred}$) and edge detection ($E_{gt},E_{pred}$). It is beneficial to localize what to preserve and enhance the semantic awareness of the network. Hence, the mapping function can be expressed as:
 	\begin{equation}
 		O=\varphi(I,I_{N},I_{P},S_{gt},S_{pred},E_{gt},E_{pred})
 	\end{equation}
 
 	We propose a Contrastive Semantic-guided Image Smoothing Network (CSGIS-Net) to preserve salient structures while removing insignificant details. Fig. \ref{net1} exhibits the training process of CSGIS-Net and demonstrates that the proposed network is able to remove the synthetic texture, even if the input images are meaningless in the real world. That is, CSGIS-Net is able to learn to distinguish the representations between unnecessary textures (e.g., bricks) and meaningful semantic structures. Hence, the model is well generalized to real-world scenes and deals with challenging textures. As demonstrated in Fig. \ref{net1}, CSGIS-Net contains an image smoothing network, a contrastive learning strategy and a segmentation network. Specifically, we employ JESS \cite{feng2021easy2hard} as our smoothing network and produce the result by bridging the edge detection. For the feature extractor, we adopt the VGG16 \cite{simonyan2014very} pre-trained on ImageNet \cite{deng2009imagenet}. For the semantic segmentation module, Deeplabv3+ \cite{chen2018encoder} is employed.

 	\begin{figure*}[!htb]
 		\centering
 		\includegraphics[width=1\textwidth]{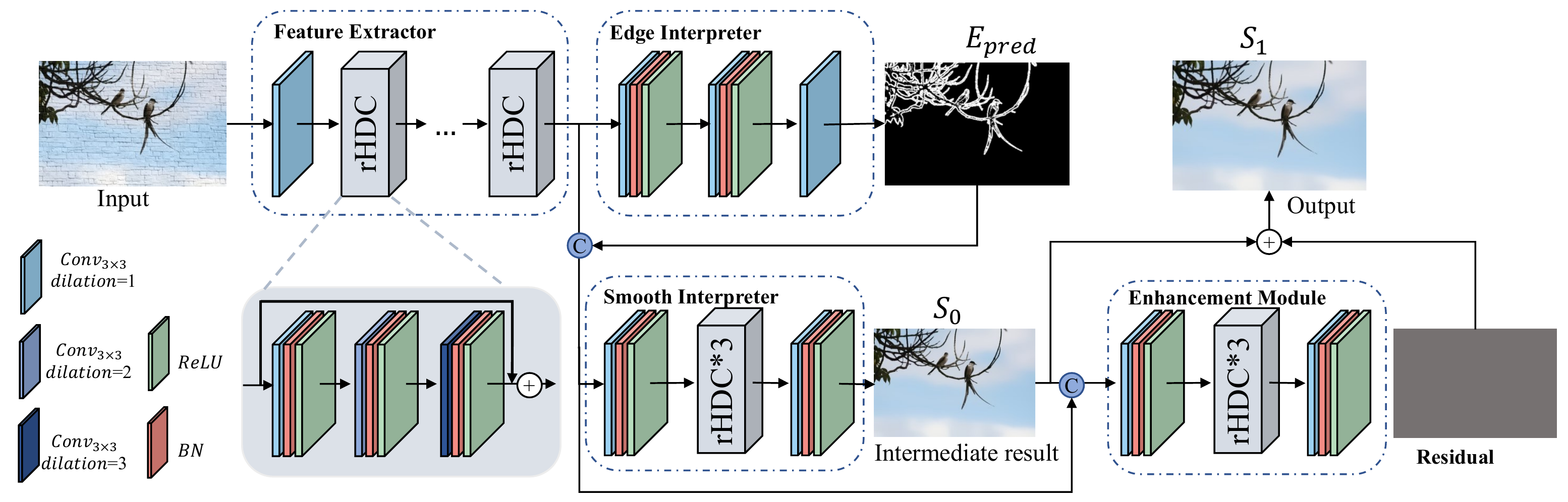}			
 		\caption{The architecture of the smoothing network.}
 		\label{net2}
 	\end{figure*}

 	\subsection{Image Smoothing Network}

 Our smoothing network comprises a feature extractor, two interpreters (i.e., image smoothing and edge detection), and a feature enhancement module inspired by \cite{deng2020detail} (see Fig. \ref{net2}). Given an image $I$, we first apply the feature extractor to extract the image-level feature maps. After that, the extracted features are fed into the edge interpreter for edge prediction, which is formulated as:
 	\begin{equation}
 		\begin{aligned}
 			F=feature\_extractor(I) \\
 			E_{pred}=edge\_interpreter(F)
 		\end{aligned}		
 	\end{equation}
 	where the feature extractor consists of a convolutional layer and series residual hybrid dilated convolution(rHDC) blocks \cite{feng2021easy2hard}. Note that, the ground-truth edge map generated from smoothing is regarded as semantic knowledge. Then the smooth interpreter receives the edge prediction $E_{pred}$ and the global features $F$ and fuses them in a concatenation way, it can be expressed as:
 	\begin{equation}
 		S_0=smooth\_interpreter(F||E_{pred})
 	\end{equation}
 	where $smooth\_interpreter$ denotes rHDC operations, $||$ denotes the concatenation. The outputs of the aforementioned modules will be delivered to the enhancement module, which is constructed by stacking multiple rHDC blocks. The final result is obtained as:
 	\begin{equation}
 		S_1=S_0 + enhance\_module(F||E_pred||S_0)
 	\end{equation}
 	where $S_0$ denotes the intermediate result and $S_1$ denotes the final smoothing prediction.
 
 The rHDC block uses the dilated convolution \cite{7913730} in a cascaded way. It gains a dense receptive field to learn the high-level image semantics, which highly improves the performance. In addition, the two interpreters perform in both parallel and cascaded manners, thereby the output of the edge interpreter will affect the parameter updating process of the smooth interpreter. Furthermore, the intermediate network parameters of the edge interpreter would not be ignored.
 	
The loss function consists of an edge detection term, a smoothing term, and a regularization term. For the edge term, it is formulated via the weighted cross entropy:
 	\begin{equation}
 		\begin{aligned}			
 			L_{e d g e}\left(E_{\text {pred }}, E_{g t}\right)&=-\alpha \sum_{i \in E^{+}} \log (p)-(1-\alpha) \sum_{i \in E^{-}} \log (1-p) \\
 			\alpha&=\frac{\left|E^{-}\right|}{|E|}, p=\operatorname{Pr}\left(e_{j}=1 \mid E\right) 			
 		\end{aligned}
 	\end{equation}
 	where $E^+(E^-)$ denotes the (non-)edge pixel set. And the smoothing term can be written as:
 	\begin{equation}
 		L_{r e}\left(S, S_{g t}\right)=L_{1}\left(S, S_{g t}\right)+\sum_{i} \operatorname{SSIM}\left(P_{2^{i}}\left(S\right), P_{2^{i}}\left(S_{g t}\right)\right)
 	\end{equation}
 	where $P_{n}$ denotes the max pooling with $n \times n$ aiming to measure the SSIM in a multi-scale manner. The edge-guided regularization term is expressed as:
 	\begin{equation}
 		L_{D T V}\left(S, S_{g t}\right)=L_{1}\left(S, S_{g t}, E^{+}\right)+T V\left(S, S_{g t}, E^{-}\right)
 	\end{equation}
 	The TV loss encourages smoothness in the non-edge domain, and the L1 term contributes to the structure consistency.
 	
 	\subsection{Contrastive Constraint}
 	Prior networks supervised by limited smoothing ground-truths often result in the domain shift and low generalization towards real-world domain. They neglect the over/under-smoothed images, such extra supervision would help the network to perceive negative representations and meet a performance gain. Inspired by the success of contrastive learning in low-level vision tasks, which aims to learn a representation that pulls the predicted image together with positive samples and pushes it far away from the negative ones, we tend to augment the supervision by fully exploring the information behind the negative samples. Therefore, three problems are proposed: 1) How to define the positive/negative samples? 2) How to construct a latent feature space? 3) How do measure the distance in the feature space? 
 	
\begin{figure}[!htb]
		\centering
		\includegraphics[width=0.45\textwidth]{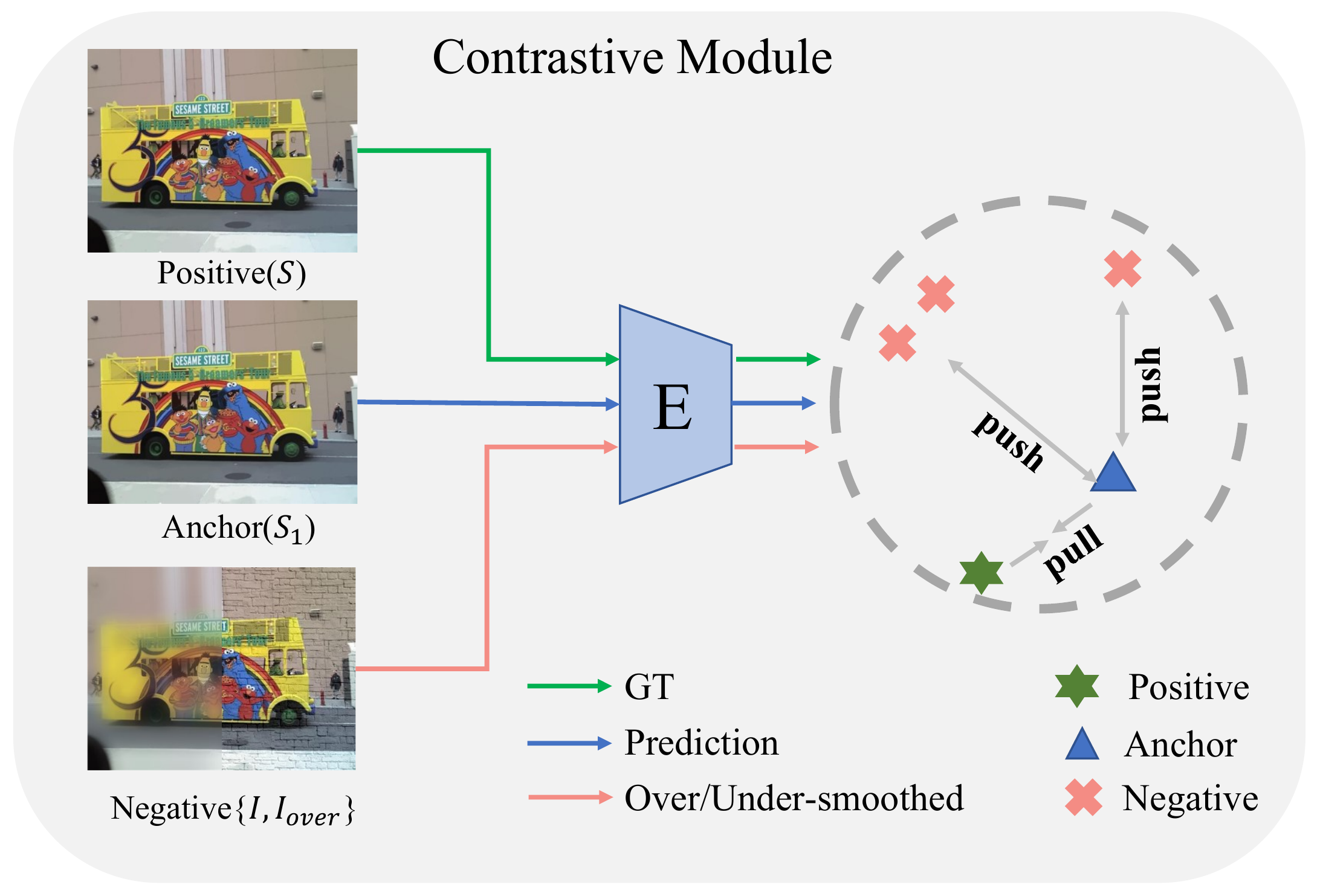}	\caption{The architecture of the contrastive module.}
		\label{net3}
	\end{figure}

 	In our work, we present the contrastive module to facilitate the training process, see from Fig. \ref{net3}. For structure preservation, we argue that the over-smoothed images usually lose the semantic structure and suffer from artifacts and halos. For the image smoothing, an input without any operation is full of complicated textures, which is opposite to our target. Therefore, as illustrated in Fig. \ref{net3}, we regard the ground-truth $S_{gt}$ as a positive sample, the original input $I$ and over-smoothed image $I_{over}$ as negative ones, and the prediction $S_1$ as an anchor. That is, $I_P=\{S\}$, $I_N=\{I,I_{over}\}$. We then adopt the pretrained VGG16 to construct the latent space. Due to the high sparsity of the deep feature space, we measure the distance based on the channel correlation. Technically, for the feature map of a convolutional layer $f \in \mathbb{R}^{C \times H \times W}$, a channel-wise Gram matrix $G\in \mathbb{R}^{C \times C}$, $G_{i j}^{l}=\sum_{k} f_{i k}^{l} f_{j k}^{l}$ is obtained firstly. In a sense, the Gram matrix represents the relationship between channels: the diagonal elements focus on self-correlation while others tend to reveal cross-correlations, which is a quantitative description of the latent feature. Therefore, our goal can be formulated as follows:
 	\begin{equation}
 		\begin{aligned}
 			d\left(G\left(\widetilde{S_{1}}\right), G\left(\widetilde{I_{P}}\right)\right) \ll d\left(G\left(\widetilde{S_{1}}\right), G\left(\widetilde{I_{N}}\right)\right) \\
 			d\left(E\left(\widetilde{S_{1}}\right), E\left(\widetilde{I}_{P}\right)\right) \ll d\left(E\left(\widetilde{S_{1}}\right), E\left(\widetilde{I_{N}}\right)\right)		
 		\end{aligned}
 	\end{equation}
 	where $S_1,I_P,I_N$ denote the anchor, positive sample, and negative sample respectively. ${(\widetilde\cdot)}$ denotes the feature map in VGG16, $E(\cdot)$ denotes the expectation. Inspired by the triplet loss, our loss function can be formulated as follows:
 	\begin{equation}
 		\begin{aligned}
 			L_{c} &=\max \left\{d\left(G\left(\breve{S}_{1}\right), G\left(\breve{I}_{P}\right)\right)-d\left(G\left(\breve{S}_{1}\right), G\left(\breve{I}_{N}\right)\right)+\alpha, 0\right\} \\
 			&+\max \left\{d\left(E\left(\breve{S}_{1}\right), E\left(\breve{I}_{P}\right)\right)-d\left(E\left(\breve{S_{1}}\right), E\left(\breve{I_{N}}\right)\right)+\beta, 0\right\}
 		\end{aligned}
 	\end{equation}

 	\subsection{Semantic Structure Preservation}
 	Although the introduction of the contrastive module has improved the performance of the network, smoothing images from a semantic perspective is still a challenge since strong details (non-semantic) and weak yet semantic structures are often intertwined in real data.
 	
 	To further improve the model's capacity for structure retention and encourage semantic distinctiveness, we introduce a semantic segmentation module to explore the high-level semantic information to benefit our CSGIS-Net. In fact, experiments have demonstrated that well-smoothed images can boost the performance of semantic segmentation tasks. That is, if the image is filled with non-semantic textures, the performance of the segmentation will be significantly dropped. It means that applying semantic supervision would meet a performance gain. Additionally, for non-synthetic images, since textures vary greatly across different object semantics, segmentation maps would guide the smoothing network to be semantic adaptive. We consider employing the mainstream network Deeplabv3+ \cite{chen2018encoder} to perform the segmentation task and adopt the cross entropy loss to ensure the consistency at the semantic level. The loss function can be defined as:
 	\begin{equation}
 		L_{seg} = CrossEntropy(\phi(S_1), Seg_{gt})
 	\end{equation}
 	where $\phi(\cdot)$ denotes the mapping of Deeplabv3+, $Seg_{gt}$ denotes the segmentation labels of PASCAL VOC 2012.
 	
 	Generally, the total loss is represented as follows:
 	\begin{equation}
 		\begin{aligned}
 			L_{\text {total }} =\lambda_{e} L_{e d g e}\left(E_{p r e d}, E_{g t}\right)+ L_{r e}\left(S_{0}, S_{g t}\right)+L_{r e}\left(S_{1}, S_{g t}\right) \\
 			+ L_{D T V}\left(S_{0}, S_{g t}\right)+ L_{D T V}\left(S_{1}, S_{g t}\right)+\lambda_{c} L_{c}+\lambda_{s e g} L_{s e g}
 		\end{aligned}
 	\end{equation}
 
 	We set $\lambda_{e}=0.001, \lambda_{c}=\lambda_{seg}=1$.

 	\section{Experiments}
 	In this section, we first establish a dataset with smoothing, edge, and segmentation ground-truths for subsequent experiments. Then, we evaluate the proposed method against several smoothing methods on two synthetic datasets quantitatively and qualitatively. Finally, several ablation studies are conducted to verify the key contribution of our method and reveal how the established ground-truth domain affects the smoothing results.

	\subsection{Dataset}\label{pro}
	
	Inspired by \cite{xu2020pixel}, structure-preserve smoothing can be regarded as a structure restoration task. However, it is difficult to determine the ground-truths of nature images. To our knowledge, most existing datasets on detection and segmentation (e.g., Cityscapes \cite{cordts2016cityscapes}, MS COCO \cite{lin2014microsoft}, PASCAL VOC 2012 \cite{everingham2015pascal}) lack the smoothing ground-truth, which is necessary for our supervised method. To realize the designed network, a synthetic dataset is established that first bridges image smoothing and semantic segmentation tasks together. It contains smoothing ground-truths and segmentation labels, where the former is also used to generate the texture-enhanced images (i.e input of the network). For pixel-wise segmentation labels, since the manual annotation is extremely laborious, we build the dataset based on the PASCAL VOC 2012 benchmark \cite{everingham2015pascal}. After that, our main task can be converted into two subtasks: 1) How to generate the smoothing ground-truth images? 2) How to design the blending strategy to generate the input?
	
	\textbf{Ground-truths Generation:} For the ground-truths of semantic smoothing, the following properties should be satisfied: 1) no severe distortion; 2) be salient in semantic structure domain; 3) smooth out non-semantic textures. 
	
	Previous methods usually fail to perform well since most of them do not consider the distinctive needed in structure and non-structure areas. Intuitively, these methods will smooth "easy" images (salient structure and small details) with better efficiency. Hence, we first collect a set of images from PASCAL VOC 2012, which is relatively easy to handle. More than 800 images are selected roughly, and we feed them into different smoothing operators such as BF \cite{tomasi1998bilateral}, L0 \cite{xu2011image}, RGF \cite{zhang2014rolling}, RTV \cite{xu2012structure}, SDF \cite{ham2017robust} to generate a series of candidates. Due to the duality of our task, that is structure preservation and detail smoothing, the evaluations of smoothness and structure similarity are performed in different regions.
	
	\emph{Smoothness Evaluation:} We argue that a visual-pleasing smoothed patch should have the following properties: 1) no extreme or sharp color oscillations (such as edges); 2) the transition of the pixel value is relatively smooth (e.g. no strong noise). Technically, we first set a window with a specific size, and slide it to search the region without edges. Note that, the edge label is obtained through a Sobel operator. After that, previous work has proved that the 1st derivative(gradient) of an image would reflect the variance rate of the gray level, and the 2nd derivative is more sensitive to the noise. To perform patch-level measurements, we combine the 2nd derivative and the standard variance, which can be formulated as follows:
	\begin{equation}
		\begin{aligned}
			L(x, y)=\left(\partial^{2} I / \partial x^{2}\right)+\left(\partial^{2} I / \partial y^{2}\right) \\
			smooth\_value_i = std(patch_i) + L(x,y)_i
		\end{aligned}
	\end{equation}
	Where $patch_i \in [0,1]$ denotes the gray value of the $i-th$ patch. Experiments show that it is nearly close to zero. Hence, we set 0.05 as a threshold value, the candidate would pass the smooth test if the smooth value is less than the threshold.
	
	\emph{Structure Consistency Evaluation:} The non-smoothed patches are obtained via a voting strategy (i.e. more than 90\% of the volunteers agree that the patch contains semantic structure). Once a volunteer reports a non-ignorable distortion, the corresponding image would be directly abandoned. It is notable that volunteers only make binary judgments, which is very simple and fast. Their judgment is only used for excluding ineligible candidates rather than being the evidence for final decisions as in \cite{zhu2019benchmark}. To measure the structural integrity, we adopt the SSIM \cite{wang2004image} formulated as:
	\begin{equation}
		SSIM(x, y)=[l(x, y)]^{\alpha} \times[c(x, y)]^{\beta} \times[s(x, y)]^{\gamma} 	
	\end{equation}
	where $l(x, y)=\frac{2 \mu_{x}  \mu_{y}+c_{1}}{\mu_{x}^{2}+\mu_{y}^{2}+c_{1}}$, $c(x, y)=\frac{2 \sigma_{x} \sigma_{y}+c_{2}}{\sigma_{x}^{2}+\sigma_{y}^{2}+c_{2}}$, $s(x, y)=\frac{\sigma_{x y}+c_{3}}{\sigma_{x} \sigma_{y}+c_{3}}$ represent brightness, contrast and structure. $\alpha$, $\beta$, $\gamma$ represent the importance of three items respectively, $\mu_i$, $\sigma_{i}$, $\sigma_{xy}$ denote the mean value, standard deviation, and covariance of the gray images respectively. The candidate passing the smooth test with the largest SSIM value will be the smooth ground-truth.
	
	Generally, we select $N$ easy samples and 5 smoothing operators, the smoothing result of the $i^{th}$ sample is represented as $O_i=\left\{f_{1}\left(I_{i}\right), f_{2}\left(I_{i}\right), \cdots, f_{p}\left(I_{i}\right)\right\}$, where $f_{i}(\cdot)$ denotes the $i_{th}$ smoothing operation. The final ground-truth images obtained by the screening test can be represented as $F=\{S_1, S_2, \cdot, S_n\}$, where $S_{i} \in O_{j}, n \leq N$. Note that, no more than one ground-truth for a single sample.
	
	\textbf{Texture Blending:} To remove texture details and preserve semantic structures, based on the well-smoothed ground-truth, we naturally introduce the texture blending process and apply the algorithm presented by \cite{efros2001image}. Such a strategy easily distinguishes the texture and structure of an input image. Technically speaking, we collect 100 texture images from \cite{feng2021easy2hard}, which cover different materials (such as grass, brick, fiber, etc.) and shapes (point, line, etc.). The algorithm is summarized as follows:
	
	1) Convert the texture image to grayscale:
	\begin{equation}
		I_{\text {gray }}=0.299 \times I_{R}+0.587 \times I_{G}+0.114 \times I_{B}
	\end{equation}
	
	where $I_R, I_G, I_B$ denote the pixel value in $R, G, B$ channels respectively.
	
	2) Calculate the mean value:
	\begin{equation}
		\mu=\frac{\sum_{i=0}^{H} \sum_{J=0}^{W} I_{g r a y}(i, j)}{H \times W}
	\end{equation}
	
	where $H, W$ denot  the height and weight of the input image.
	
\begin{figure}[!htb]
		\centering
		\includegraphics[width=0.45\textwidth]{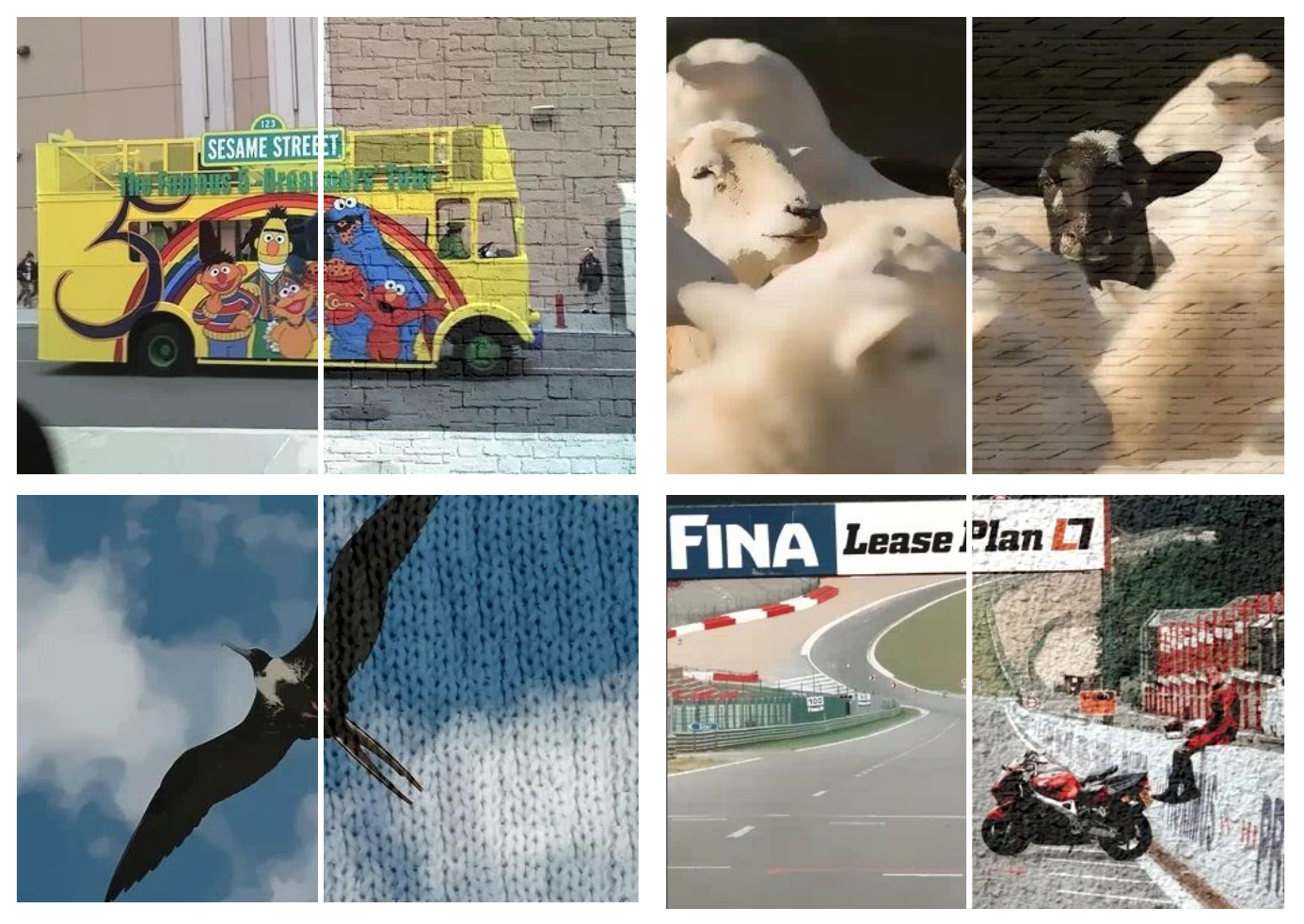}			
		\caption{Some examples of our established dataset. The left is ground-truth images, the right is the inputs of our network.}
		\label{data}
	\end{figure}

	3) Texture blending:
	\begin{equation}
		I=I_{GT}+(I_{gray}-\mu)
	\end{equation}
	
	where $(I_{gray}-\mu)$ represents the texture layer, and $I_{GT}$ is the ground-truth image. The blending result $I$ would be fed into the network as an input. Fig. \ref{data} shows some examples.
 	
 	\subsection{Implementation Details}
 	 The established VOC-smooth dataset consists of 15k images, which blends 100 texture layers and 150 smoothing ground-truths. Among them, 12k images are chosen randomly for training, 2600 images for validation, and 400 images for testing. Images for testing are not overlapped with the training images. In order to better generalize our method to various cases, we adopt a series of data augmentations: 1) random horizontal flip; 2) random image rotation; 3) color disturbances. The proposed CSGIS-Net is trained in a fully supervised fashion.

	Our networks are implemented on the Pytorch platform and trained on an NVIDIA GeForce GTX 1080Ti GPU. During the training phase, the patch size is set to $128\times 128$ and the batch size is set to 8. For accelerating the training process, the ADAM optimizer \cite{kingma2014adam} is adopted with $\beta_{1}=0.9,\beta_{2}=0.999$. The initial learning rate $1e-3$ decayed by multiplying 0.5 after every 30 epochs and we train 250 epochs in total. We set the number of rHDC blocks $N=3$ and $N=6$ in the smooth interpreter and the enhancement module, respectively.

 	\begin{table}[]
 		\caption{Average PSNR/SSIM comparison on the synthetic VOC-smooth and NKS datasets. The highest rank in comparison is highlighted in \textbf{bold}.}
 		\label{table1}
 		\centering
 		\begin{tabular}{lllll}
 			\hline
 			& \multicolumn{2}{c}{VOC-smooth}     & \multicolumn{2}{c}{NKS}            \\ \cline{2-5} 
 			& SSIM            & PSNR             & SSIM            & PSNR             \\ \hline
 			Original & 0.3882 & 20.9701  & 0.5736  & 28.0418          \\
 			GF  \cite{he2012guided}     & 0.5452  & 23.7863  & 0.8809  & 32.0883 \\
 			RGF \cite{zhang2014rolling} & 0.7076  & 22.3704  & 0.9033  & 31.0875 \\
 			L0  \cite{xu2011image}      & 0.4684  & 21.9665  & 0.7854  & 28.4811 \\
  			RTV \cite{xu2012structure}  & 0.7775  & 24.8952  & 0.9221  & 31.8124 \\
 			SDF \cite{ham2017robust}    & 0.5312  & 22.8657  & 0.9077  & 33.1725 \\
 			ILS \cite{liu2020real}      & 0.5474  & 22.3584  & 0.9050  & 29.6929 \\
			PNLS \cite{xu2020pixel}     & 0.6622  & 23.6922  & 0.9341  & 33.2073 \\
 			JESS\cite{feng2021easy2hard}& 0.8788  & 28.3701  & 0.9435  & 33.0475 \\
 			Ours     & \textbf{0.8993} & \textbf{29.3366} & \textbf{0.9486} & \textbf{34.5098} \\ \hline
 		\end{tabular}
 	\end{table}

 	\subsection{Comparison Results}
 	In the testing phase, we adopt two synthetic datasets (VOC-smooth and NKS \cite{xu2020pixel}) to quantitatively evaluate different methods. All images in the two datasets are blended by texture layers and diverse structure images (i.e. ground-truths). Note that the NKS dataset is blended by cartoon images, which is quite simple compared to real-world images. We apply the Peak Signal to Noise Ratio(PSNR) and the Structural Similarity(SSIM) for evaluation, and higher values indicate better performance.
 	
 	Our proposed method is compared with eight representative smoothing algorithms, including two filters(i.e. GF \cite{he2012guided}, RGF \cite{zhang2014rolling}), five optimization-based methods(i.e. L0 \cite{xu2011image}, RTV \cite{xu2012structure}, SDF \cite{ham2017robust}), ILS \cite{liu2020real}, PNLS \cite{xu2020pixel} and a data-driven network JESS \cite{feng2021easy2hard}. We determine the model with the best performance in the validating phase for further analysis. The results are summarized in Tab. \ref{table1}, by averaging the PSNR and SSIM of the test set. Note that, for traditional methods, we choose the best result from a set of manual parameters. It can be observed that our CSGIS-Net outperforms the others on the two synthetic datasets. Specifically, compared to the second best methods, the PSNRs of VOC-smooth and NKS are 0.9665 and 1.3373 higher, respectively. 
 	
\begin{figure*}[!htb]
 		\centering
 		\includegraphics[width=0.81\textwidth]{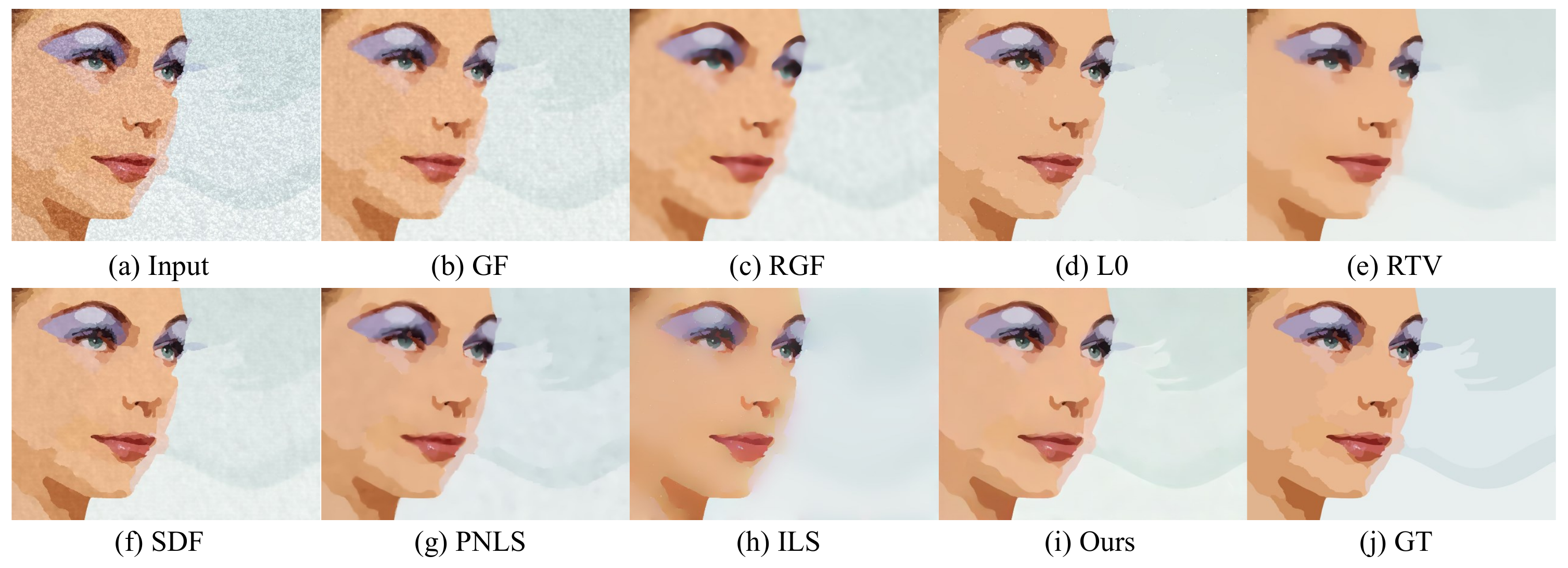}			
 		\caption{Visual Results on NKS. (b) GF ($\sigma=2.5$); (c) RGF ($\sigma_{c}=0.3, \sigma_{s}=6,iter=4$); (d) L0 ($\lambda=0.01$); (e) RTV ($\epsilon_s=0.02, iter=3$); (f) SDF ($\lambda=3$); (g) PNLS ($ thr=0.4, iter = 9$); (h) ILS ($\lambda=30, \gamma=10/255,iter=10$).}
 		\label{fig1}
 	\end{figure*}

\begin{figure*}[!htb]
 		\centering
 		\includegraphics[width=0.81\textwidth]{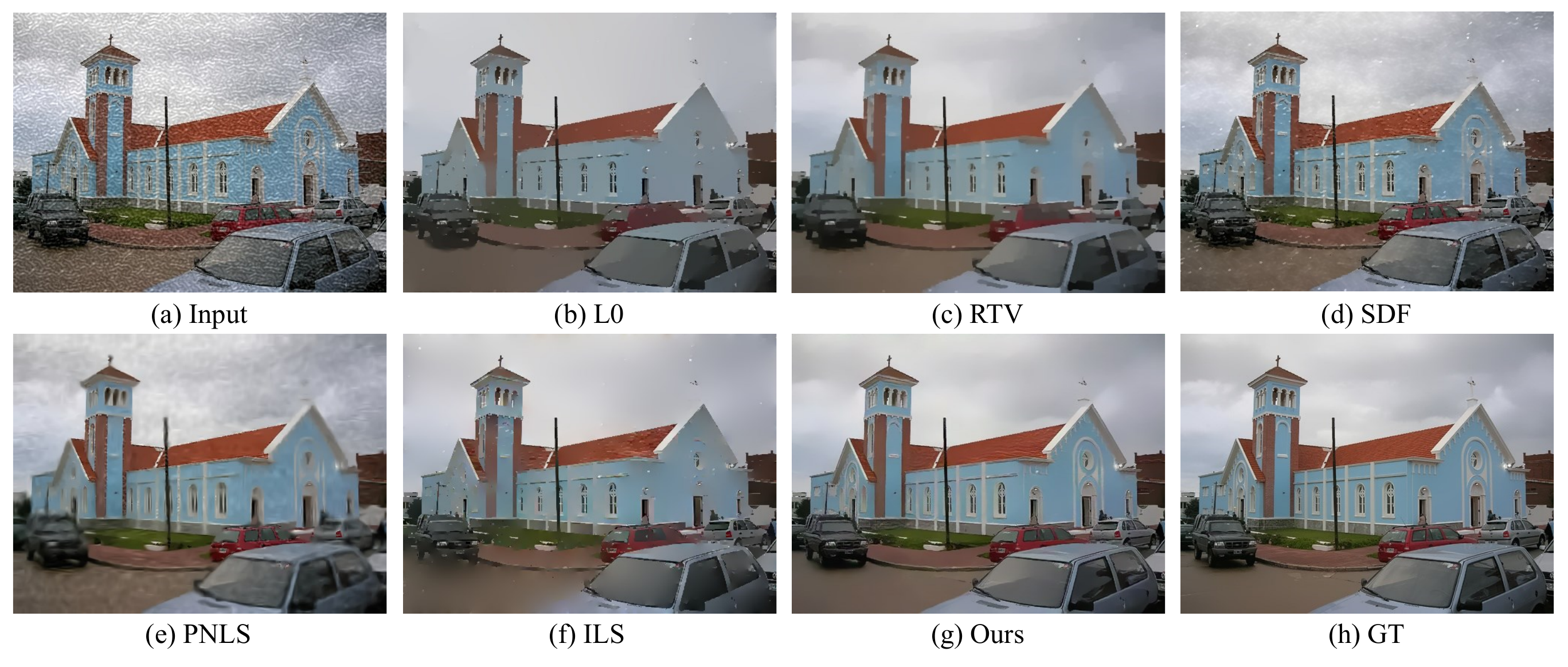}			
 		\caption{Visual Results on VOC-smooth. (b) L0 ($\lambda=0.1$); (c) RTV ($\epsilon_s=0.01, iter=2$); (d) SDF ($\lambda=3$); (e) PNLS ($ thr=0.4, iter = 9$); (f) ILS $\lambda=30, \gamma=13/255,iter=10$}
 		\label{fig2}
 	\end{figure*}

\begin{figure*}[!htb]
 		\centering
 		\includegraphics[width=0.81\textwidth]{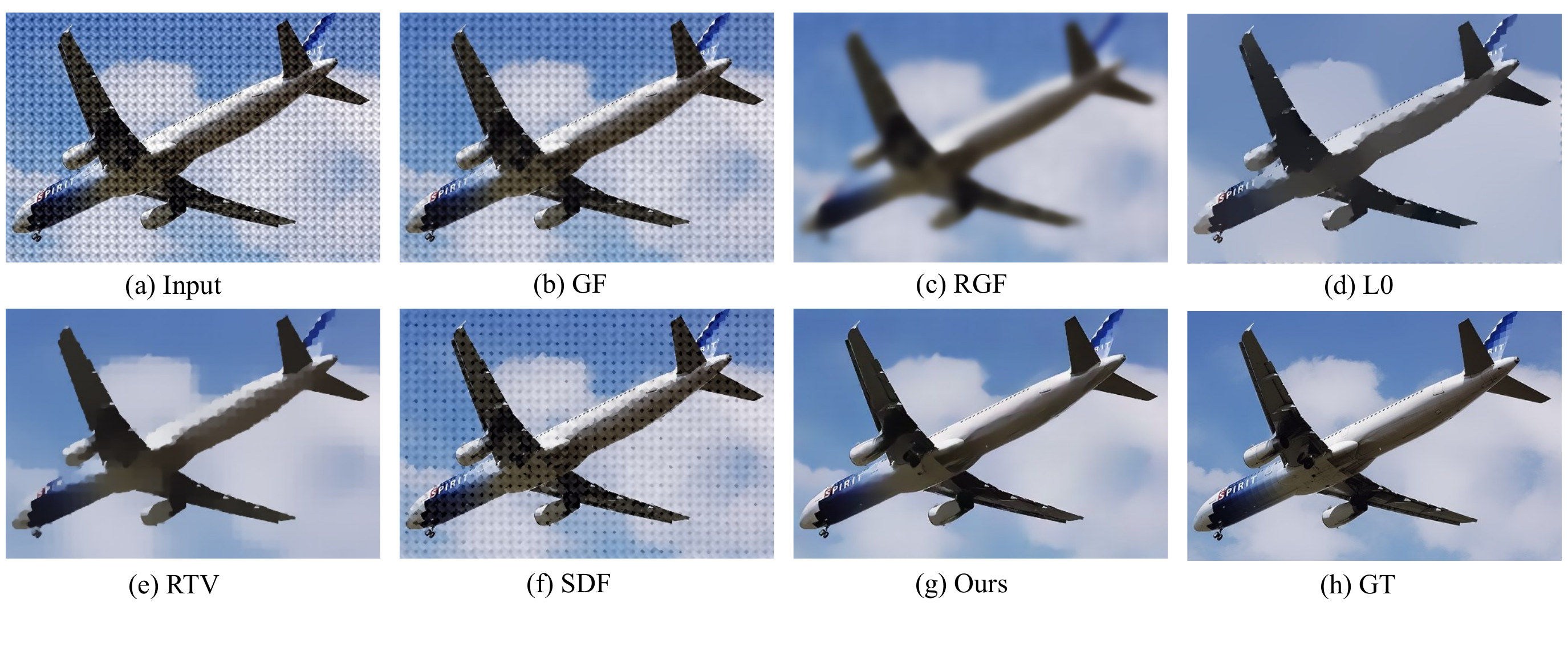}			
 		\caption{Visual Results on VOC-smooth. (b) GF ($\sigma=2.5$); (c) RGF ($\sigma_{c}=0.3, \sigma_{s}=6,iter=4$); (d) L0 ($\lambda=0.08$); (e) RTV ($\epsilon_s=0.02, iter=3$); (f) SDF ($\lambda=3$).}
 		\label{fig3}
 	\end{figure*}

 	We also exhibit visual comparisons where the parameters of traditional algorithms are manually determined via the visual quality. As shown in Fig. \ref{fig1}, \ref{fig2}, \ref{fig3}, GF and SDF can hardly remove the texture layer. L0 and RTV remove texture with appropriate parameters, but some salient structures are also overblurred, such as eyes, windows. Those methods fail to achieve a trade-off between texture removal and structure preservation since the large-scale smoothing strength will introduce distortion and overblurring, while the small-scale one is disabled to smooth the texture. However, our proposed network can achieve notable performance compared with other algorithms.
 	
 	In fact, we expect that our network would be generalized to the real domain. In the testing phase, the chosen real-world image contains strong textures or complex details, such as grass, road, etc. Since the absence of smoothing ground-truth, we only exhibit the visual comparisons of these two categories, i.e. 1) natural images; 2) real textural images. 
 
\begin{figure*}[!htb]
 		\centering
 		\includegraphics[width=1\textwidth]{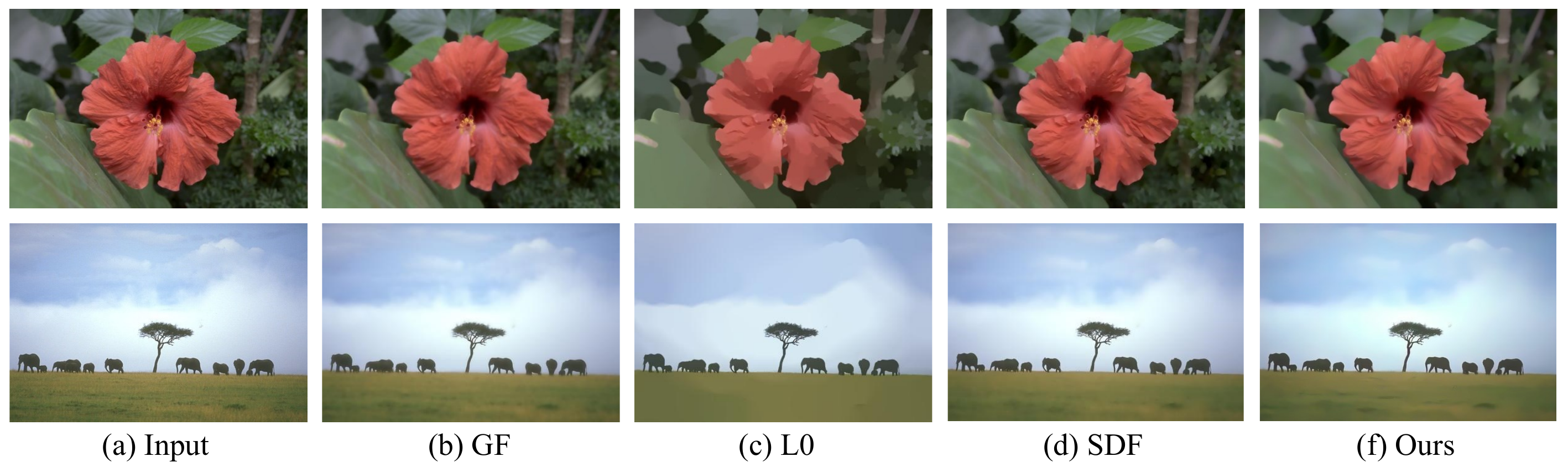}			
 		\caption{Visual results on a natural image. (b) GF ($\sigma=2.5$); (c) L0 ($\lambda=0.01$); (d) SDF ($\lambda=3$).}
 		\label{fig4}
 	\end{figure*}

\begin{figure*}[!htb]
 		\centering
 		\includegraphics[width=1\textwidth]{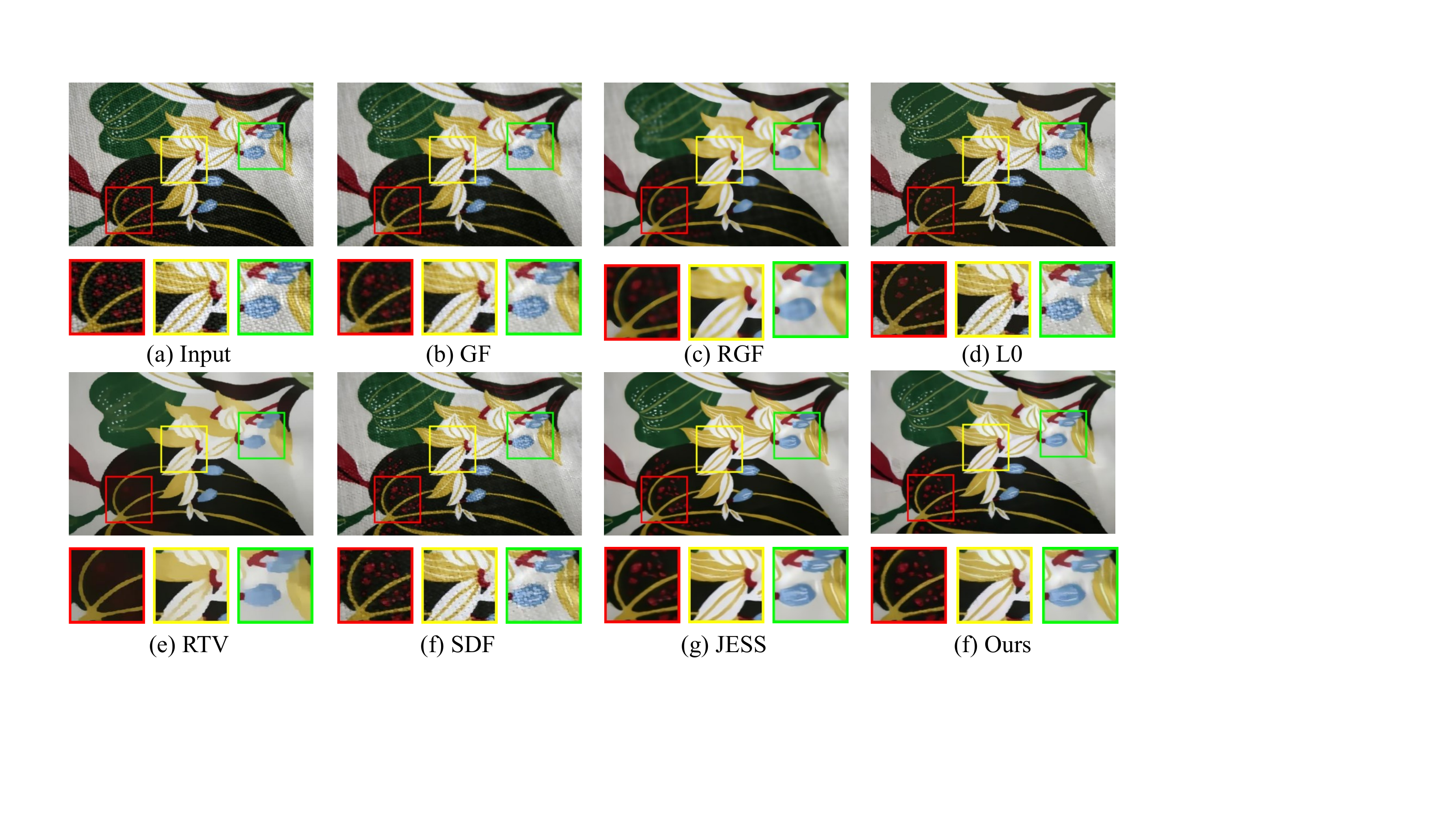}			
 		\caption{Visual results on a real texture image. (b) GF ($\sigma=2.5$); (c) RGF ($\sigma_{c}=0.3, \sigma_{s}=6,iter=4$); (d) L0 ($\lambda=0.01$); (e) RTV  ($\epsilon_s=0.02, iter=3$); (f) SDF ($\lambda=3$).}
 		\label{fig5}
 	\end{figure*}

 	As depicted in Fig. \ref{fig4},\ref{fig5}, our approach generates considerably better results while others suffer blurring or distortion to varying degrees. For the natural images, the smooth strength of GF is small, leading to the retention of small textures such as leaves; L0 loses salient edges and introduces artifacts; SDF fails to smooth the clouds. Our method produces a more natural mapping in the texture layer. For the real textual image, GF, L0, and SDF have limited ability for texture removal. RGF and RTV gain better smoothing performance but neglect semantic structures such as red dots and white lines(see from Fig. \ref{fig5}). Our proposed network possesses a better generalization ability to real data, due to the exploration of the contrastive and semantic priors.

	\begin{table*}[!htb]
		\centering 		
		\caption{Quantitative comparison for the network ablation study.}
		\label{table2}
		\begin{tabular}{ccccccc}
			\hline
			
			& \multirow{2}{*}{Contrastive Module} & \multirow{2}{*}{Semantic Module} & \multicolumn{2}{c}{VOC-smooth} & \multicolumn{2}{c}{NKS} \\ \cline{4-7} 
			&                                     &                                  & SSIM          & PSNR           & SSIM       & PSNR       \\ \hline
			CSGIS-baseline &   &    & 0.8381   & 25.6451  & 0.9276  &30.0736    \\
			CSGIS-CM  & \checkmark  &  & 0.8478  & 26.4537 & 0.9310 &30.3613    \\
			CSGIS   & \checkmark   & \checkmark   & \textbf{0.8647} & \textbf{27.4539} & \textbf{0.9340}  & \textbf{32.2314}    \\ \hline
		\end{tabular}
	\end{table*}
	
	\begin{table}[]
		\centering
		\setlength{\tabcolsep}{6mm}
		\caption{Quantitative comparison for the dataset ablation study.}
		\label{table3}
		\begin{tabular}{ccc}
			\hline
			\multirow{2}{*}{Training Dataset} & \multicolumn{2}{c}{NKS} \\ \cline{2-3} 
			& SSIM          & PSNR           \\ \hline
			Under-smoothed                        & 0.9191        & 29.2022        \\
			Over-smoothed                      & 0.9076        & 29.8602        \\
			VOC-smooth                        & \textbf{0.9340}        & \textbf{32.2314}        \\ \hline
		\end{tabular}
	\end{table}

  	\begin{figure*}[!htb]
		\centering
		\includegraphics[width=1\textwidth]{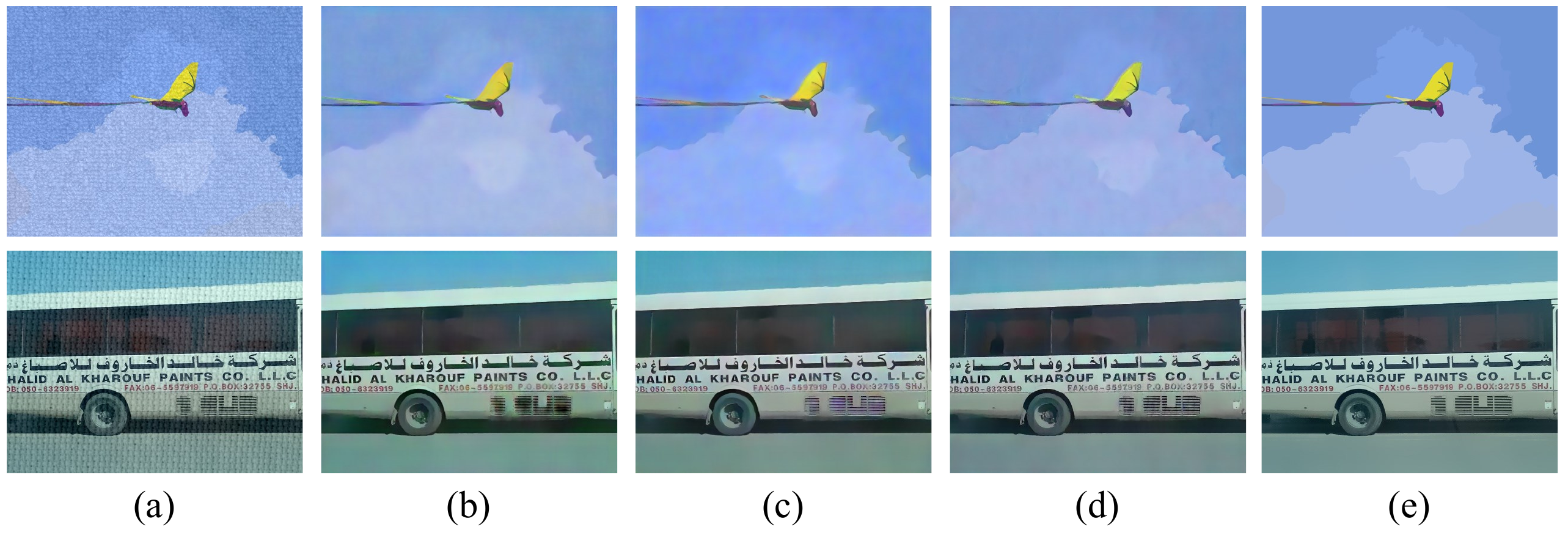}			
		\caption{Visual results for ablation study. (a) Input images; (b) Baseline, no Contrastive module (CM) and Semantic segmentation module (SM); (c)CSGIS-CM, only adding CM; (d) CSGIS-Net, the network of this paper; (e) GT, the ground-truth. The first row is from the NKS dataset and the second row is from the VOC-smooth dataset.}
		\label{fig6}
	\end{figure*}
	
	\begin{figure*}[!htb]
		\centering
		\includegraphics[width=1\textwidth]{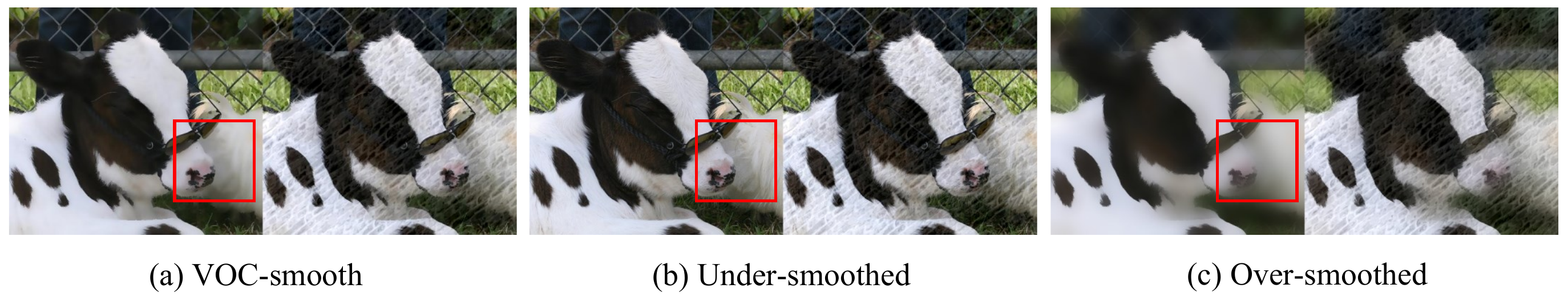}			
		\caption{The training datasets of three different domains. The left is ground-truths and the right is input images. Over/under-smoothed means the ground-truth images are over/under-smoothed.}
		\label{fig8}
	\end{figure*}

	\begin{figure*}[!htb]
		\centering
		\includegraphics[width=0.9\textwidth]{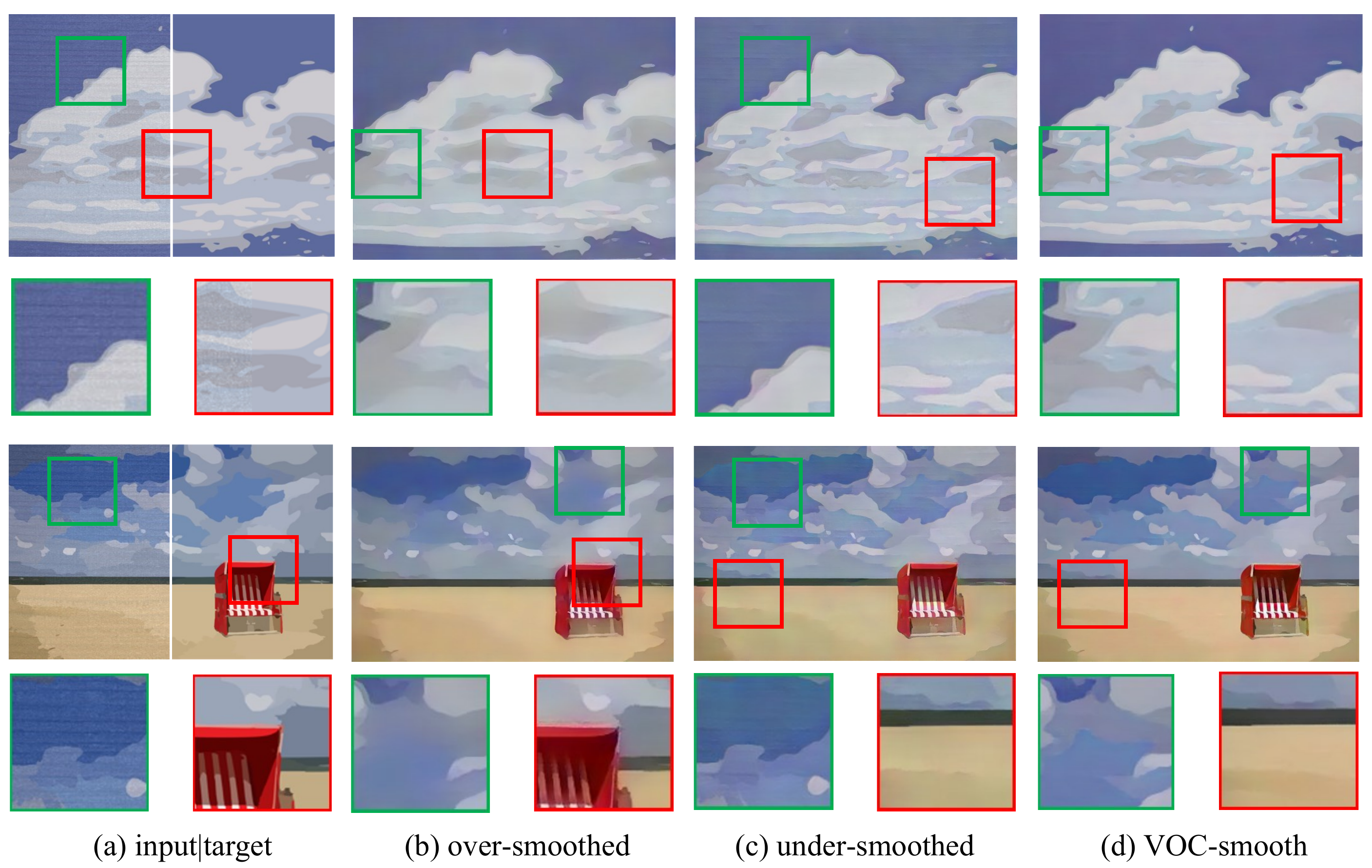}			
		\caption{Visual results for dataset ablation study.}
		\label{fig7}
	\end{figure*}

	\subsection{Ablation Study}
	In this subsection, extensive ablation studies are conducted to verify the effectiveness of the proposed components in our framework, including 1) two introduced modules of contrastive learning and image segmentation; 2) the established dataset (i.e. the training domain). Considering the training process on the entire dataset would take about 3-4 days with 250 epochs, all the ablation studies are organized on the subset of our datasets (VOC-smooth). Specifically, we randomly choose 2600 images for training, 400 images for validating and 400 images for testing. 
	
	\noindent
	\textbf{Ablation in Network Architecture}
	
	We discuss the contributions of the introduced modules and evaluate the performance on the two datasets, i.e. VOC-smooth and NKS, the quantitative comparison is illustrated in Tab. \ref{table2}. One can notice that the network with all proposed modules achieves the best performance. Specifically, compared to the backbone, the PSNR and SSIM are improved by 1.8088 and 0.0266 respectively. Moreover, simply adding a contrastive module also meets a performance gain. A typical example is shown in Fig. \ref{fig6}. It can be observed that (b) and (c) overblur some edges on the cloud and butterfly's wings (see from the first row). Furthermore, the second row demonstrates that our network generates visual-pleasing results, especially for semantic structure preservation. It is inferred that contrastive learning with negatives may prevent the prediction from the over/under-smoothed domain.
	
	Note that using only the contrastive loss to train the proposed network may not yield satisfactory results. This is due to the sparsity of the contrastive module with the high-dimensional feature space, where small changes may have a large impact on the low-dimensional smoothing results. Furthermore, our recent experiments demonstrate that the model fails to converge when only constrained by the contrastive loss, implying that a low-dimensional loss is necessary.
	
	\noindent
	\textbf{Ablation in Dataset}
	
We explore the potential impact of the established dataset (VOC-smooth). Firstly, over/under-smoothed domains are constructed separately for training. The former takes over-smoothed images as smoothing ground-truths, which are obtained by a Bilateral Filter with a large smooth strength ($\sigma_{c}=200, \sigma_{s}=1000$). The latter uses the original images from PASCAL VOC 2012 as ground-truths, it is interwoven with the real texture. We then apply the aforementioned blending strategy to synthesize the input images. Retrain our model in three different domains, for fairness, we only test on the NKS dataset quantitatively.

	As shown in Tab. \ref{table3}, the network trained on VOC-smooth gains the highest PSNR and SSIM. The visual results also illustrate that the VOS-smooth domain performs favorably against others (see Fig. \ref{fig7} ). We argue that for the over-smoothed one, the ground-truth itself loses semantic structures and clear edges. Therefore, the network fails to learn the representation of structures, leading to the problem of blurring. Moreover, the network trained on the under-smoothed domain has a limited ability in texture removal, since the mixture of synthetic and real texture confuses the network as to what to remove/preserve. However, benefiting from the three properties in sec. \ref{pro}, our dataset contributes to smoothing the non-semantic details under strong texture.
	
	\section{Conclusions}
	In this work, we propose a novel network (CSGIS-Net) that aims to preserve the salient structure of the image while smoothing the insignificant details by introducing two important prior knowledge. CSGIS-Net augments the supervision from unexplored over/under-smoothed samples and semantic information, which promote the network to be more robust and semantic adaptive. Based on the introduced contrastive module and segmentation module, the network is regarded as a multi-task framework with three parts of loss function for semantic structure preservation and texture removal. Additionally, we establish a new dataset (VOC-smooth) that mixes versatile structure images with natural textures. It also makes the first trial to bridge image smoothing and semantic segmentation. Extensive experiments demonstrate that the proposed CSGIS-Net outperforms the state-of-the-art smoothing methods both visually and quantitatively. In the future, we will extend our method to other applications such as detail enhancement, and further explore other strategies for exploiting semantic information.

\section*{Acknowledgments}
This work was supported by the National Natural Science Foundation of China (No. 62172218), the Free Exploration of Basic Research Project, Local Science and Technology Development Fund Guided by the Central Government of China (No. 2021Szvup060), the Natural Science Foundation of Guangdong Province (No. 2022A1515010170),  the Direct Grant (No. DR22A2) and the Research Grant entitled "Self-Supervised Learning for Medical Images" (No. 871228) of Lingnan University, Hong Kong.

	\bibliographystyle{eg-alpha-doi} 
	\bibliography{egbibsample}       
	
	
	\newpage
	
\end{document}